\documentclass[11pt,letterpaper]{article}
\usepackage{url}
\usepackage{caption,subcaption}
\usepackage{graphicx}
\usepackage{naaclhlt2016}
\usepackage{times}
\usepackage{latexsym}

\naaclfinalcopy 
\def\naaclpaperid{***} 

\newcommand{\mysubsection}[1]{\vspace{0.3em} \noindent\textbf{#1}}

\title{Stateology: State-Level Interactive Charting of \\Language, Feelings, and Values}

\author{Konstantinos Pappas, Steven Wilson, \and Rada Mihalcea \\
     Computer Science and Engineering\\
     University of Michigan\\
     Ann Arbor, MI 48109\\
     \{pappus, steverw, mihalcea\}@umich.edu\\}

\date{}

\begin{document}

\maketitle

\begin{abstract}
People's personality and motivations are manifest in their everyday language usage. With the emergence of social media, ample examples of such usage are procurable. In this paper, we aim to analyze the vocabulary used by close to 200,000 Blogger users in the U.S. with the purpose of geographically portraying various demographic, linguistic, and psychological dimensions at the state level. We give a description of a web-based tool for viewing maps that depict various characteristics of the social media users as derived from this large blog dataset of over two billion words.
\end{abstract}

\section{Introduction}

Blogging gained momentum in 1999 and became especially popular after the launch of freely available, hosted platforms such as \url{blogger.com} or  \url{livejournal.com}. Blogging has progressively been used by individuals to share news, ideas, and information, but it has also developed a mainstream role to the extent that it is being used by political consultants and news services as a tool for outreach and opinion forming as well as by businesses as a marketing tool to promote products and services \cite{Nardi04}.

For this paper, we compiled a very large geolocated collection of blogs, written by individuals located in the U.S., with the purpose of creating insightful mappings of the blogging community. In particular, during May-July 2015, we gathered the profile information for  all  the users  that  have  self-reported  their  location in  the U.S., along with a number of posts for all their associated blogs. We utilize this blog collection to generate maps of the U.S. that reflect user demographics, language use, and distributions of psycholinguistic and semantic word classes. We believe that these maps can provide valuable insights and partial verification of previous claims in support of research in linguistic geography \cite{brice03}, regional personality \cite{rogers10}, and language analysis ~\cite{schwartz13,schmitt07}, as well as psychology and its relation to human geography \cite{kitchin97}.

\begin{figure*}[t]
\begin{minipage}[c][6cm][t]{.5\textwidth}
  \vspace*{\fill}
  \centering
  \includegraphics[width=8.39cm,height=4.7cm]{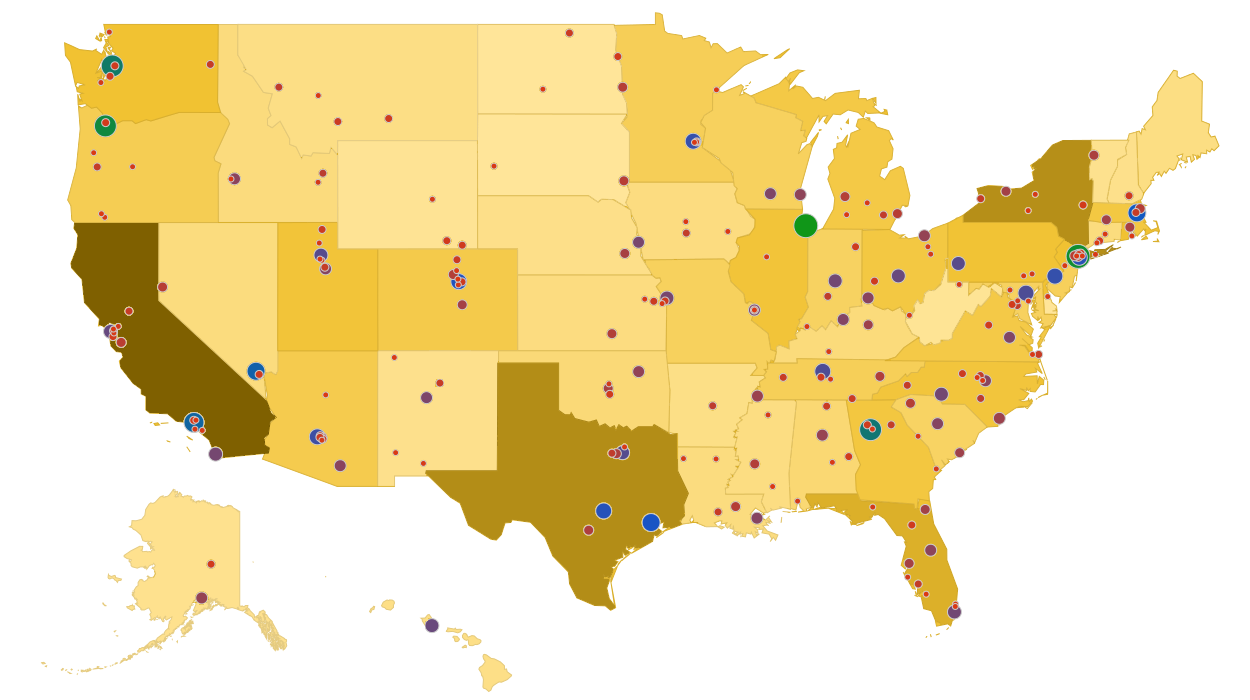}
  \subcaption{States and cities density}
  \label{fig:statecity}
\end{minipage}%
\begin{minipage}[c][6cm][t]{.5\textwidth}
  \vspace*{\fill}
  \centering
  \includegraphics[width=3.56cm,height=2.2cm]{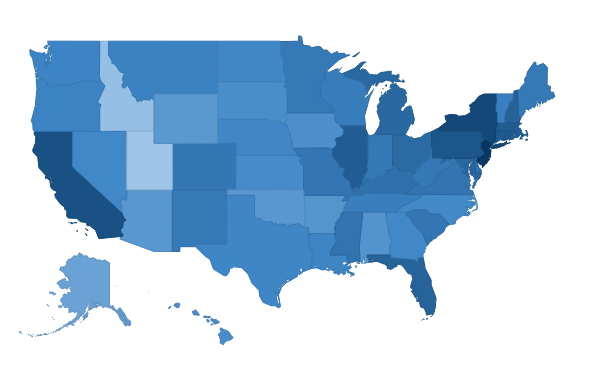}
  \subcaption{Male}
  \label{fig:male}\par\vfill
  \includegraphics[width=3.56cm,height=2.2cm]{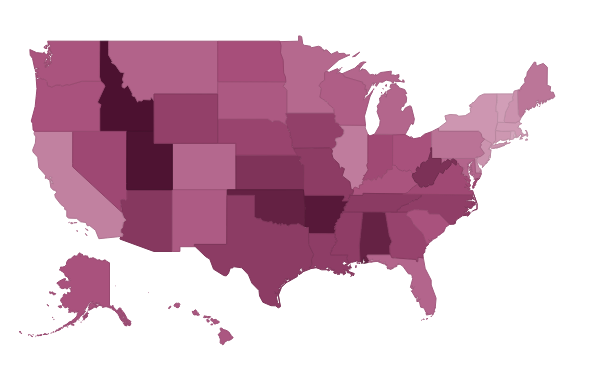}
  \subcaption{Female}
  \label{fig:female}
\end{minipage}
\caption{Geographical distribution of bloggers in the 50 U.S. states.\label{figure:distMap}}
\end{figure*}

\section{Data Collection}
Our premise is that we can generate informative maps using geolocated information available on social media; therefore, we guide the blog collection process with the constraint that we only accept blogs that have specific location information. Moreover,  we aim to find blogs belonging to writers from all 50 U.S. states, which will allow us to build U.S. maps for various dimensions of interest.

We first started by collecting a set of profiles of bloggers that met our location specifications by searching individual states on the profile finder on \url{http://www.blogger.com}. Starting with this list, we can locate the profile page for a user, and subsequently extract additional information, which includes fields such as name, email, occupation, industry, and so forth. It is important to note that the profile finder only identifies users that have an exact match to the location specified in the query; we thus built and ran queries that used both state abbreviations (e.g., TX, AL), as well as the states' full names (e.g., Texas, Alabama).

After completing all the processing steps, we identified 197,527 bloggers with state location information. For each of these bloggers, we found their blogs (note that a blogger can have multiple blogs), for a total of 335,698 blogs. For each of these blogs, we downloaded the 21 most recent blog postings, which were cleaned of HTML tags and tokenized, resulting in a collection of 4,600,465 blog posts.

\begin{figure*}
\small
\begin{tabular*}{1\textwidth}{@{\extracolsep{\fill}} c c }
\multicolumn{1}{c}{\includegraphics[width=0.45\textwidth]{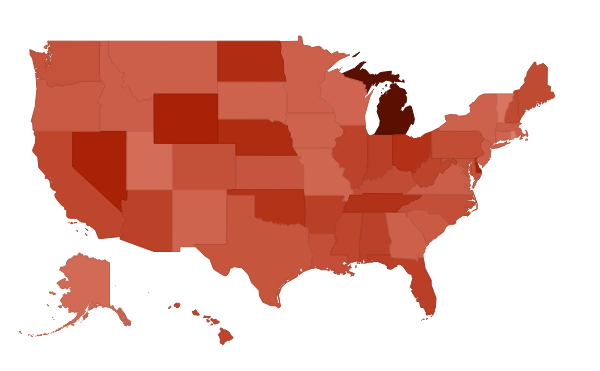}} &\multicolumn{1}{c}{\includegraphics[width=0.45\textwidth]{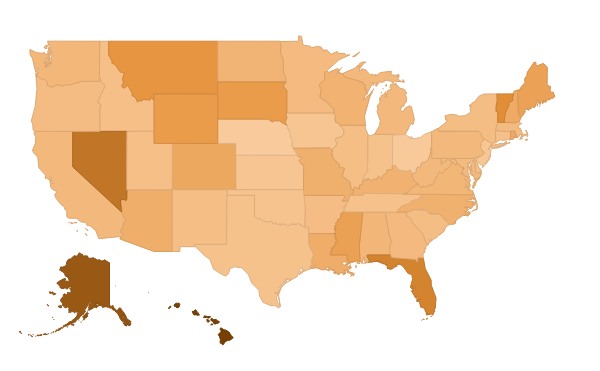}}\\
\multicolumn{1}{c}{Automotive} & \multicolumn{1}{c}{Tourism}\\
\end{tabular*}
\caption{Industry distribution across the 50 U.S. states for two selected industries.} 
\label{figure:industryDist}
\end{figure*} 

\begin{figure*}
\small
\begin{tabular*}{1\textwidth}{@{\extracolsep{\fill}} c c }
\multicolumn{1}{c}{\includegraphics[width=0.45\textwidth]{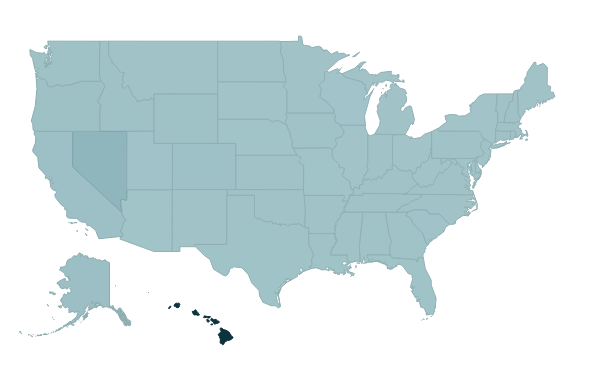}} &\multicolumn{1}{c}{\includegraphics[width=0.45\textwidth]{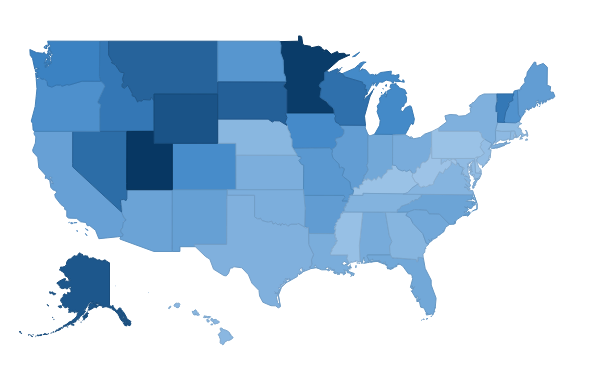}}\\
\multicolumn{1}{c}{Maui} & \multicolumn{1}{c}{Lake}\\
\end{tabular*}
\caption{Word distributions across the 50 U.S. states for two selected words.} 
\label{figure:wordsDist}
\end{figure*} 

\section{Maps from Blogs}

Our dataset provides mappings between location, profile information, and language use, which we can leverage to generate maps that reflect demographic, linguistic, and psycholinguistic properties of the population represented in the dataset.\footnote{In all the maps we generate, the darker the color of a state, the higher the proportion of instances in that state that match the criterion used to generate the map. }

\subsection{People Maps}

The first map we generate depicts the distribution of the bloggers in our dataset across the U.S. Figure \ref{fig:statecity} shows the density of users in our dataset in each of the 50 states.  For instance, the densest state was found to be California with 11,701 users. The second densest is Texas, with 9,252 users, followed by New York, with 9,136. The state with the fewest bloggers is Delaware with 1,217 users. Not surprisingly, this distribution correlates well with the population of these states,\footnote{http://www.census.gov/2010census/data/apportionment-dens-text.php} with a Spearman's rank correlation $\rho$ of 0.91 and a p-value $<$ 0.0001, and is very similar to the one reported in Lin and Halavais \shortcite{Lin04}.

Figure \ref{fig:statecity} also shows the cities mentioned most often in our dataset. In particular, it illustrates all 227 cities that have at least 100 bloggers. The bigger the dot on the map, the larger the number of users found in that city. The five top blogger-dense cities, in order, are: Chicago, New York, Portland, Seattle, and Atlanta.

We also generate two maps that delineate the gender distribution in the dataset. Overall, the blogging world seems to be dominated by females: out of 153,209 users who self-reported their gender, only 52,725 are men and 100,484 are women. Figures \ref{fig:male} and \ref{fig:female} show the percentage of male and female bloggers in each of the 50 states. As seen in this figure, there are more than the average number of male bloggers in states such as California and New York, whereas Utah and Idaho have a higher percentage of women bloggers.

Another profile element that can lead to interesting maps is the {\it Industry} field \cite{holmes04}. Using this field, we created different maps that plot the geographical distribution of industries across the country. As an example, Figure \ref{figure:industryDist} shows the percentage of the users in each state working in the automotive and tourism industries respectively.

\subsection{Linguistic Maps}

Another use of the information found in our dataset is to build linguistic maps, which reflect the geographic lexical variation across the 50 states \cite{Eisenstein10}. We generate maps that represent the relative frequency by which a word occurs in the different states. Figure \ref{figure:wordsDist} shows sample maps created for two different words. The figure shows the map generated for one location specific word, {\it Maui}, which unsurprisingly is found predominantly in Hawaii, and a map for a more common word, {\it lake}, which has a high occurrence rate in Minnesota (Land of 10,000 Lakes) and Utah (home of the Great Salt Lake). Our demo described in Section \ref{sec:demo}, can also be used to generate maps for function words, which can be very telling regarding people's personality \cite{chung07}.

\begin{figure*}
\small
\begin{tabular*}{1\textwidth}{@{\extracolsep{\fill}} c c }
\multicolumn{1}{c}{\includegraphics[width=0.45\textwidth]{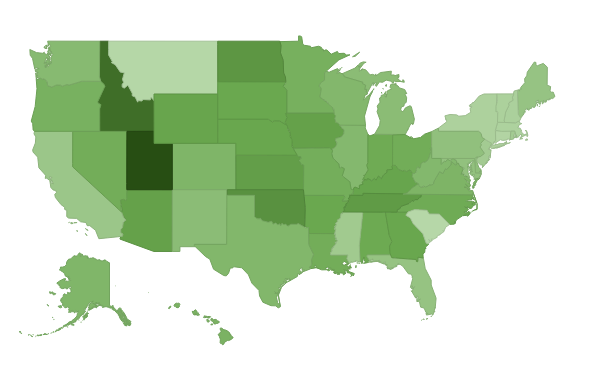}} &\multicolumn{1}{c}{\includegraphics[width=0.45\textwidth]{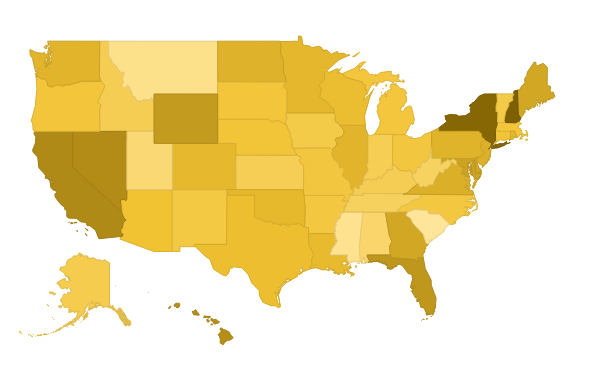}}\\
\multicolumn{1}{c}{Positive feelings} & \multicolumn{1}{c}{Money}\\
\end{tabular*}
\caption{LIWC distributions across the 50 U.S. states for two selected semantic categories.} 
\label{figure:liwcDist}
\end{figure*} 

\begin{figure*}
\small
\begin{tabular*}{1\textwidth}{@{\extracolsep{\fill}} c c }
\multicolumn{1}{c}{\includegraphics[width=0.45\textwidth]{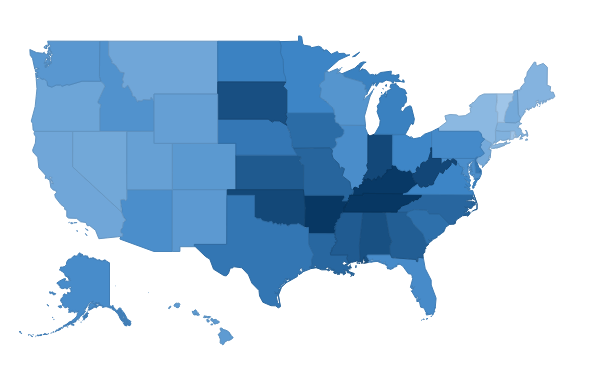}} &\multicolumn{1}{c}{\includegraphics[width=0.45\textwidth]{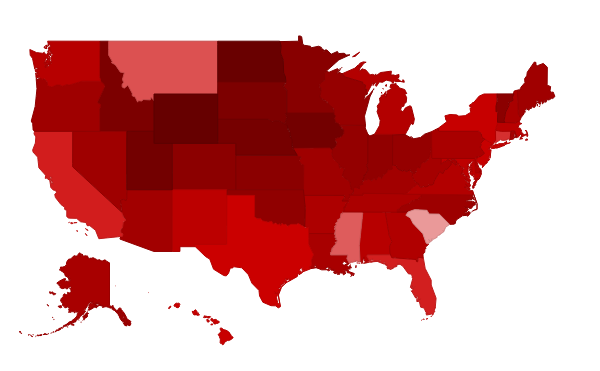}}\\
\multicolumn{1}{c}{Religion} & \multicolumn{1}{c}{Hard work}\\
\end{tabular*}
\caption{Values distributions across the 50 U.S. states for two selected values.} 
\label{figure:valuesDist}
\end{figure*}

\subsection{Psycholinguistic and Semantic Maps}

\mysubsection{LIWC.} In addition to individual words, we can also create maps for word categories that reflect a certain psycholinguistic or semantic property. Several lexical resources, such as Roget or Linguistic Inquiry and Word Count \cite{pennebaker01}, group words into categories. Examples of such categories are {\sc Money}, which includes words such as remuneration, dollar, and payment; or {\sc Positive feelings} with words such as happy, cheerful, and celebration.
Using the distribution of the individual words in a category, we can compile distributions for the entire category, and therefore generate maps for these word categories. For instance, figure \ref{figure:liwcDist} shows the maps created for two categories: {\sc Positive Feelings} and {\sc Money}. The maps are not surprising, and interestingly they also reflect an inverse correlation between {\sc Money } and {\sc Positive Feelings }. 

\mysubsection{Values.} We also measure the usage of words related to people's core values as reported by Boyd et al. \shortcite{boyd2015}. The sets of words, or themes, were excavated using the Meaning Extraction Method (MEM) \cite{chung2008}. MEM is a topic modeling approach applied to a corpus of texts created by hundreds of survey respondents from the U.S. who were asked to freely write about their personal values. To illustrate, Figure \ref{figure:valuesDist} shows the geographical distributions of two of these value themes: {\sc Religion} and {\sc Hard Work}. Southeastern states often considered as the nation's ``Bible Belt'' \cite{heatwole1978} were found to have generally higher usage of {\sc Religion} words such as {\it God}, {\it bible}, and {\it church}. Another broad trend was that western-central states (e.g., Wyoming, Nebraska, Iowa) commonly blogged about {\sc Hard Work}, using words such as {\it hard}, {\it work}, and {\it job} more often than bloggers in other regions.

\section{Web Demonstration}
\label{sec:demo}

A prototype, interactive charting demo is available at \url{http://lit.eecs.umich.edu/~geoliwc/}. In addition to drawing maps of the geographical distributions on the different LIWC categories, the tool can report the three most and least correlated LIWC categories in the U.S.\footnote{We use the Spearman rank correlation coefficient to calculate correlation.} and compare the distributions of any two categories.

\section{Conclusions}

In this paper, we showed how we can effectively leverage a prodigious blog dataset. Not only does the dataset bring out the extensive linguistic content reflected in the blog posts, but also includes location information and rich metadata. These data allow for the generation of maps that reflect the demographics of the population, variations in language use, and differences in psycholinguistic and semantic categories. These mappings can be valuable to both psychologists and linguists, as well as lexicographers. A prototype demo has been made available together with the code used to collect our dataset.\footnote{\url{http://lit.eecs.umich.edu/downloads.html}}

\section*{Acknowledgments}

This material is based in part upon work supported by the National Science Foundation (\#1344257) and by the John Templeton Foundation (\#48503). Any opinions, findings, and conclusions or recommendations expressed in this material are those of the authors and do not necessarily reflect the views of the National Science Foundation or the John Templeton Foundation. We would like to thank our colleagues Hengjing Wang, Jiatao Fan, Xinghai Zhang, and Po-Jung Huang who provided technical help with the implementation of the demo.

\bibliographystyle{naaclhlt2016}

\bibliography{naaclhlt2016}

\end{document}